# Augmenting Biomedical Named Entity Recognition with General-domain Resources

Yu Yin[1,*], Hyunjae Kim[2,*], Xiao Xiao[1], Chih Hsuan Wei[3], Jaewoo Kang[2], Zhiyong Lu[3], Hua Xu[4], Meng Fang[1,†], Qingyu Chen[4,†]

1. Department of Computer Science, University of Liverpool, Liverpool L69 3DR, United Kingdom
2. Department of Computer Science, Korea University, 145 Anam-ro, Seongbuk-gu, Seoul, 02841, Republic of Korea
3. National Center for Biotechnology Information, National Library of Medicine, National Institutes of Health, Bethesda, Maryland, 0894, United States of America
4. Department of Biomedical Informatics and Data Science, School of Medicine, Yale University, New Haven, Connecticut, 06510, United States of America

*These authors contributed equally to this work.

+Corresponding authors: Meng.Fang@liverpool.ac.uk, qingyu.chen@yale.edu.

## Abstract:

**Objective:** Training a neural network-based biomedical named entity recognition[1] (BioNER) model usually requires extensive and costly human annotations. While several studies have employed multi-task learning with multiple BioNER datasets to reduce human effort, this approach does not consistently yield performance improvements and may introduce label ambiguity in different biomedical corpora. We aim to tackle those challenges through transfer learning from easily accessible resources with fewer concept overlaps with biomedical datasets.

**Methods:** We proposed GERBERA, a simple-yet-effective method that utilized general-domain NER datasets for training. We performed multi-task learning to train a pre-trained biomedical language model with both the target BioNER dataset and the general-domain dataset. Subsequently, we fine-tuned the models specifically for the BioNER dataset.

**Results:** We systematically evaluated GERBERA on five datasets of eight entity types, collectively consisting of 81,410 instances. Despite using fewer biomedical resources, our models demonstrated superior performance compared to baseline models trained with additional BioNER datasets. Specifically, our models consistently outperformed the baseline models in six out of eight entity types, achieving an average improvement of 0.9% over the best baseline performance across eight entities. Our method was especially effective in amplifying performance on BioNER datasets characterized by limited data, with a 4.7% improvement in F1 scores on the JNLPBA-RNA dataset.

**Conclusion:** This study introduces a new training method that leverages cost-effective general-domain NER datasets to augment BioNER models. This approach significantly improves BioNER model performance, making it a valuable asset for scenarios with scarce or costly biomedical datasets. We make data, codes, and models publicly available via https://github.com/qingyu-qc/bioner_gerbera.

---

[1] **Abbreviations,** BioNER, Biomedical Named Entity Recognition.

## Graphical Abstract :

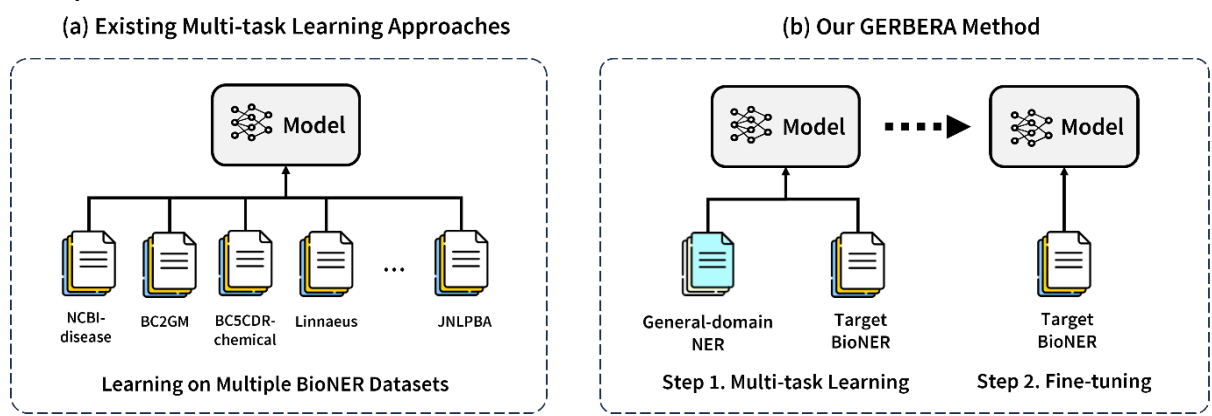

## 1. Introduction

Named entity recognition (NER) is a fundamental task in natural language processing (NLP) that focuses on identifying and classifying named entities such as persons, organizations, and locations within text. The biomedical named entity recognition (BioNER) task focuses on identifying biomedical entity types such as diseases, genes, proteins, and chemicals within the input text. It is commonly used for downstream applications such as question answering [1-3], relation extraction [4-7], and biomedical knowledge graph construction [8-10]. With the surge in the amount of biomedical literature, information overload has become a serious problem [11], which intensifies the need for automatic BioNER techniques. Moreover, the accuracy of BioNER also has a critical impact on the quality of downstream applications.

In recent years, various pre-trained biomedical language models have emerged as popular approaches for BioNER tasks. They can be categorized into (1) encoder-based, using the encoder from the Transformer [12] architecture, such as BioBERT, BlueBERT, and PubMedBERT [13-17]; (2) decoder-based, using the decoder from the Transformer architecture, such as BioGPT and BioMedLM [18, 19]; and (3) encoder-decoder-based, using both encoders and decoders, such as BioBART and SciFive [20, 21]. Despite the emergence of large language models (LLMs), domain-specific biomedical language models still remain the state-of-the-art (SOTA) in the field of biomedical research [22-24], with encoder-based models in particular holding the forefront.

However, a primary challenge in building BioNER models using the SOTA approach is their heavy reliance on manually-annotated gold standard corpora for model training. Obtaining large-scale annotated BioNER datasets is often expensive and challenging because manual annotation by biomedical experts requires a high-level of domain expertise and is time-consuming [25, 26]. To address the issue of insufficient training data and improve model performance, several studies have proposed multi-task learning approaches for BioNER [28-31]. These approaches involve training a single BioNER model simultaneously using multiple BioNER datasets with different types of entity annotations. For example, when training a disease NER model, datasets for the "gene," "chemical," and "species" types can also be used in the training process. Nevertheless, applying multi-task learning to BioNER tasks presents two significant limitations: 1) it does not consistently achieve optimal performance across all

BioNER datasets simultaneously as different tasks are hard to get converged at the same time point; 2) utilizing multiple BioNER datasets for training could introduce label ambiguity stemming from overlapping concepts across different biomedical corpora. To mitigate this, the intervention of domain experts is often necessary to meticulously review and reconcile differences between the various BioNER datasets incorporated into the training process [32, 33].

Figure 1: The performance of multi-task learning models across multiple BioNER datasets.

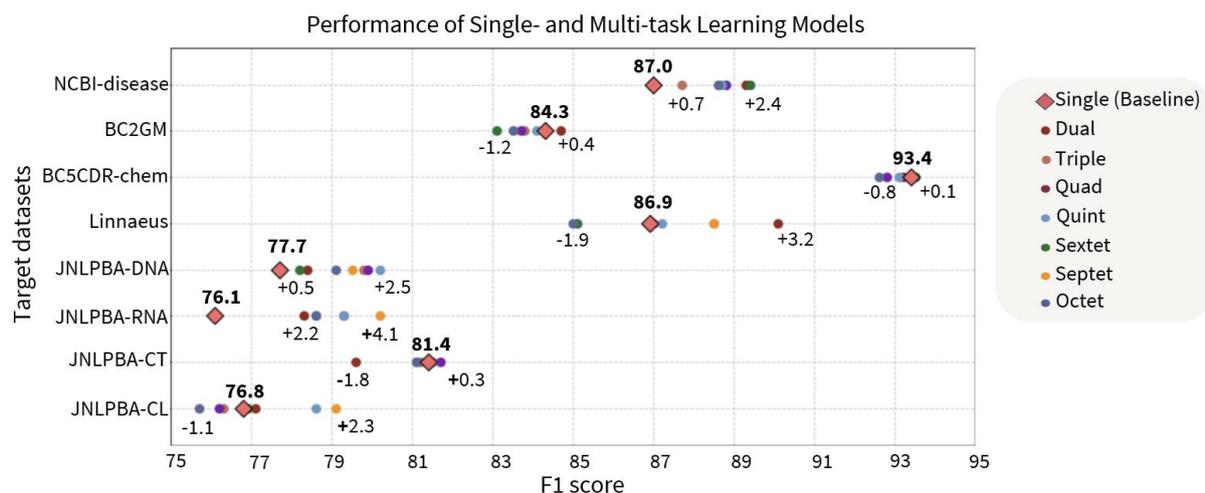

*Note*: The legend entries from 'Single' to 'Octet' represent the number of BioNER datasets included during training. The 'Single' model serves as the baseline for comparison with multi-task learning models. Baseline performances are highlighted in bold, with annotations indicating the maximum and minimum performance deviations from the baseline.

To delve into the difficulties of multi-task learning in BioNER, we evaluated the performance of a multi-task learning model across various biomedical datasets. We randomly sampled the following BioNER datasets and added them one by one to train the model: NCBI-disease [34], BC2GM [35], BC5CDR-chemical [36], Linnaeus [37], JNLPBA-DNA, JNLPBA-RNA, JNLPBA-CT, and JNLPBA-CL [27].[2] The "Single" model was trained on the target dataset alone, while the "Double" to "Octet" models were trained with additional randomly sampled datasets. Figure *1* demonstrates that multi-task learning does not consistently improve performance. For instance, while the "Octet" model, trained with eight datasets, showed improved performance on some datasets, it performed worse on others. Notably, it achieved the lowest scores across all experimental settings on the Linnaeus and JNLPBA-CL datasets.[3]

We hypothesize that augmenting BioNER models with general-domain NER datasets can address those challenges, inspired by knowledge transfer in general domains [38, 39]. First, they encompass a wider range of entity types beyond the biomedical domain, minimizing the likelihood of label ambiguity when trained together with BioNER datasets. Second, compared with BioNER datasets which are mostly derived from biomedical literature, general-domain NER datasets introduce more diverse linguistic features, potentially aiding in the generalization capability of NER models. Moreover, general-domain NER datasets–in contrast to resource-intensive biomedical-domain annotations–are much more accessible with significantly larger data scales and annotation depth. Building upon these insights, we propose GERBERA, short for *Transfer Learning for **Ge**neral-to-**B**iomedical **E**ntity **R**ecognition **A**ugmentation*, as a way to improve the performance of the model on BioNER datasets. Specifically, our proposed transfer learning method is based on multi-task learning that leverages knowledge from general-domain NER datasets to improve performance on BioNER datasets, without introducing new biomedical resources.

---

[2] The various BioNER datasets are further introduced in Section 2.1. Additionally, Table 1 provides a detailed summary of the statistics for each dataset used in our study.
[3] This could be due to label ambiguity caused by overlapping concepts across different corpora. For instance, entities such as "monocytic", "transfected", and "osteosarcoma" in the JNLPBA-CL dataset overlap with concepts in the NCBI-disease, Linnaeus, and JNLPBA-CT corpus.

Our experiments demonstrate that combining individual general-domain NER datasets with individual BioNER datasets can significantly enhance the ability of the model to recognize boundary and missing cases, thereby improving the overall F1 value performance of the model, especially on small-scale BioNER datasets. On average, our method increases F1 scores by 0.9% compared to baselines, with a notable improvement of 4.7% in the F1 score on the JNLPBA-RNA dataset. Our method leverages the extensive linguistic features of general-domain NER datasets to complement the highly specialized knowledge inherent in BioNER datasets, which enhances the ability of the BioNER model to more fully understand and deal with the intricate features specific to the biomedical domain. Our proposed method has demonstrated promising results on six out of eight biomedical entity types, highlighting its potential to alleviate the negative impact of data deficiency in BioNER tasks.

Our key contributions can be summarized as follows:

1. We introduce GERBERA, a simple-yet-effective training method that can leverage general-domain NER datasets to augment the BioNER task. To the best of our knowledge, we are the first to employ transfer learning using general-domain NER data in the BioNER field.
2. We thoroughly evaluated the effectiveness of our method using the five datasets with eight biomedical entity types and compared with five baselines. Our method consistently outperformed the baselines in three out of five datasets and six out of eight entity types – improving the F1 score of strong baseline performance by 0.9%.
3. Our method is efficient because it does not use resource-intensive biomedical-domain annotations. Without relying on additional BioNER datasets, it significantly improves model performance on data-limited datasets, such as the JNLPBA-RNA dataset, with a 4.7% improvement in F1 scores.
4. We make our data, codes, and models publicly available.

## 2. Materials and methods

We first review in detail the datasets used in our research, as well as the baseline models against which our method is compared. Next, we describe how to use the transfer learning approach to improve our BioNER models. Finally, we present the implementation details.

## 2.1. Datasets

As BERN2 [24] serves as our strong baseline model, we have chosen to employ a similar selection of BioNER datasets to facilitate direct comparisons. For BioNER datasets, we employ NCBI-disease [34], BC2GM [35], BC5CDR-chemical[4] [36], Linnaeus [37], and JNLPBA [27] datasets for disease, gene/protein, chemical, species, and DNA, RNA, cell line and cell type annotations, respectively. [5] Additionally, we incorporate four well-established general-domain NER datasets: CoNLL2003 [42], Gum [43], MIT_Movie,[6] and MIT_Restaurant.[7] The CoNLL2003 dataset is well-known in the field of NER tasks, which annotates four types of named entities: person (PER), organization (ORG), location (LOC), and miscellaneous (MISC). The Gum dataset is an open-source corpus including multiple entity types such as object, plant, and time. The MIT_Movie dataset is specifically designed for NER tasks related to movie queries, containing entity types such as actor names, movie titles, and genres. The MIT_Restaurant dataset is curated for recognizing entity types in the context of restaurant reviews such as rating, dish, and amenity. Statistics for datasets are shown in Table 1.

## 2.2. Baselines

The selection of the baseline models is based on well-established SOTA BioNER models from recent years. We conduct a comparative analysis of our results against the following baselines (see Table 2 for a summary): 1) **BERN2** [24], a SOTA multi-task BioNER tool based on Bio-LM [17] that can recognize multiple biomedical named entities and support entity normalization. Note that we conducted a direct comparison with BERN2 as we employ the same datasets and backbone model. 2) **AIONER** [45], a newly developed BioNER model built upon PubMedBERT [14], employed a flexible tagging scheme that integrates eleven BioNER training datasets into a unified resource, matching or surpassing the performance of previous SOTA methods, 3) **PTC** [44], a web-based system for recognizing biomedical named entity types in articles, 4) **BioBERT** [13], a pre-trained BERT based model [47] that is further pre-trained using biomedical literature including PubMed abstracts and PMC full text articles.

---

[4] BC5CDR-chemical: Referred to as BC5CDR-chem in all tables.
[5] In our future work, we plan to extend our evaluations to other widely used BioNER datasets such as S1000 [40], BC5CDR-disease [36], and BioRED [41] .
[6] https://groups.csail.mit.edu/sls/downloads/movie/
[7] https://groups.csail.mit.edu/sls/downloads/restaurant/

Table 1: Statistics of datasets used in our method.

| Dataset | Type | #Sentences (#Entities) | | |
| --- | --- | --- | --- | --- |
| | | Train | Dev | Test |
| ***PubMed Abstracts*** | | | | |
| NCBI-disease | Disease | 5432 (*5145*) | 923 (*787*) | 942 (*960*) |
| BC2GM | Gene/Protein | 12632 (*37301*) | 2531 (*7498*) | 5065 (*15101*) |
| BC5CDR-chem | Chemical | 4560 (*5203*) | 4581 (*5347*) | 4797 (*5385*) |
| Linnaeus | Species | 12004 (*3259*) | 4086 (*1064*) | 7181 (*2232*) |
| JNLPBA-DNA | DNA | 4699 (*22549*) | 552 (*2758*) | 622 (*2845*) |
| JNLPBA-RNA | RNA | 721 (*2192*) | 89 (*289*) | 102 (*305*) |
| JNLPBA-CT | Cell type | 4792 (*14324*) | 420 (*1142*) | 1422 (*4912*) |
| JNLPBA-CL | Cell line | 2596 (*10049*) | 284 (*1168*) | 377 (*1489*) |
| ***News*** | | | | |
| CoNLL2003 | Location, Person, Organization, Miscellaneous | 14041 (*23499*) | 3250 (*5942*) | 3453 (*5648*) |
| ***Movie query*** | | | | |
| MIT_Moive | Actor, Title, Director, Genre, Song, Trailer, Review, Year, Character, Ratings_average, Rating, Plot | 9775 (*21295*) | N/A | 2443 (*5339*) |
| ***Restaurant query*** | | | | |
| MIT_Restaurant | Rating, Dish, Amenity, Location, Cuisine Restaurant_Name, Hours, Price | 7660 (*15363*) | N/A | 1521 (*3151*) |
| ***Various*** | | | | |
| Gum | Abstract, Animal, Event, Object, Organization, Person, Place, Plant, Quantity, Substance, Time | 2495 (*8285*) | N/A | 1000 (*3439*) |

*Note*: All biomedical datasets are sourced from PubMed abstracts. The CoNLL2003 dataset originates from News. The MIT_Movie dataset comes from movie-related queries. The MIT_Restaurant dataset is derived from restaurant reviews. The Gum dataset comprises content from various sources, including Wikipedia, Reddit, and YouTube.

Table 2: Summary of BioNER model baselines.

| Model | Training Method |
|---|---|
| BERN2 [24] | Bio-LM large model fine-tuned using multi-task learning on eight BioNER datasets. |
| AIONER [45] | PubMedBERT base model with a CRF layer trained using eleven BioNER datasets with a unified tagging scheme. |
| PTC [44] | Biomedical text mining tool that integrates multiple entity taggers trained on PubMed corpora for BioNER tasks. |
| BioBERT [13] | BERT base model pre-trained on PubMed and PMC and fine-tuned on a single target biomedical dataset. |
| GERBERA (**Ours**) | Bio-LM large model fine-tuned using multi-task learning on one general-domain NER dataset and one target BioNER dataset. |

*Note:* BERN2 [24] and AIONER [45] utilize eight and eleven BioNER datasets for training, respectively. Our GERBERA model incorporates an additional general-domain NER dataset, significantly reducing its reliance on domain-specific datasets.

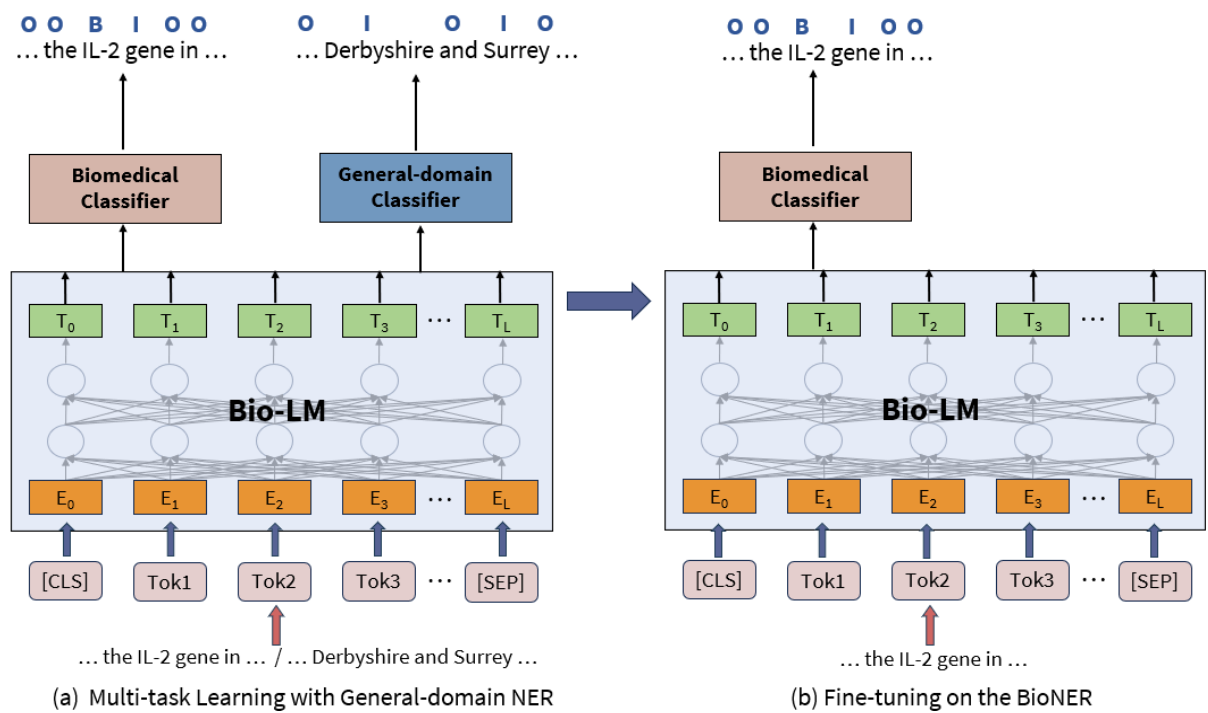

Figure 2: Our proposed method merges a general-domain dataset with a biomedical dataset as model inputs, employing Bio-LM [17] as the shared backbone, with biomedical and general NER task-specific classifiers. Biomedical fine-tuning is subsequently performed with the target BioNER dataset.

## 2.3. Transfer learning for BioNER

In transfer learning, a model is initially trained on source data ($S$) and then adapted or assessed on target data ($T$). We modify this approach by integrating multi-task learning, utilizing both source and target data simultaneously, with target data used solely for fine-tuning purposes. The target dataset is a biomedical dataset, such as JNLPBA [27], while the source dataset is a general-domain dataset, such as CoNLL2003 [42]. Our approach integrates knowledge transferred from the general-domain NER dataset and inherent biomedical insights from the BioNER dataset.

Our method leverages Bio-LM [8] as a shared backbone model, which is a pre-trained Transformer-encoder-based language model known for its strong performance in BioNER tasks [17, 45]. Furthermore, we incorporate a dual-layer multi-layer perceptron (MLP) with ReLU activation that serves as the task-specific layer including BioNER and general-domain NER classifiers. In the training phase, we start by merging the target biomedical NER dataset with a general-domain NER dataset. Next, the input merged datasets are used for multi-task learning with a cross-entropy loss. We subsequently fine-tune the model using data from the target BioNER dataset. Finally, the model outputs the probabilities of the three classes of the input sequence: 'B', 'I', and 'O'. The core objective of this integrated training approach is to assimilate valuable insights from the broad domain of general-domain NER into the more specialized realm of BioNER, thereby enhancing the richness of the BioNER model. Subsequent fine-tuning of the model on the target BioNER dataset is intended to optimize model performance by keeping the model focused on the highly specialized linguistic features of the target biomedical domain. The model architecture is shown in Figure 2.

The loss function $L$ is calculated as follows:

$$L(\theta) = -\frac{1}{N_t}\sum_{i=1}^{N_t} log\left(p(y_i^t | X_i^t, \theta)\right) - \frac{1}{N_s}\sum_{j=1}^{N_s} log\left(p\left(y_j^s | X_j^s, \theta\right)\right) \quad (1)$$

The loss function $L(\theta)$ computes the average negative log-likelihood loss across two datasets. The input text sequence from the source general-domain NER dataset is represented by $X^s$, and the input text sequence from the target BioNER dataset is represented by $X^t$. The trainable parameters are represented by $\theta$. The function $P(y_i^t | X_i^t, \theta)$ calculates the probability of correctly predicting the label $y_i^t$ at each position $i$ in the biomedical dataset, given the input sequence $X_i^t$. Similarly, $P(y_j^s | X_j^s, \theta)$ refers to the probability of correctly predicting the label $y_j^s$ in the general-domain dataset, given the input sequence $X_j^s$. The sequence length from source and target datasets are represented by $N_s$ and $N_t$, respectively. We use the above formula to fine-tune our model. For BioNER model training, rather than merging all BioNER datasets, we combine each general-domain NER dataset with each of the eight BioNER datasets.

### 2.4. Implementation Details
We pre-processed four general-domain NER datasets to ensure uniformity in the annotation format and re-annotated the relevant named entities with the 'B' and 'I' tags, while labeling all others with 'O'.[9] Additionally, each BioNER dataset was then combined with each general-domain NER dataset for co-training, hyperparameter tuning, and evaluation. For model training configurations, the learning rate was set at 3e-5, the batch size at 16, and the maximum sequence length at 128. Models were trained for a maximum of 10 epochs. Furthermore, model performance was evaluated using precision, recall, and the F1 score. During hyperparameter tuning, we adjusted the training epochs to 5, 10, 15, or 20, and varied the learning rates at 1e-5, 3e-5, or 5e-5. During the validation phase, we compared the training loss with the validation loss to identify and eliminate hyperparameter combinations that resulted in significant overfitting, specifically where the training loss was substantially lower than the validation loss. Subsequently, we selected the model with the optimal set of hyperparameters based on the highest F1 scores obtained on the validation set. Notably, the reported models were trained exclusively on the training set.

---

[8] https://github.com/facebookresearch/bio-lm
[9] Our decision to re-annotate named entities with 'B' and 'I' tags, rather than 'B-{entity type}' and 'I-{entity type}', stems from initial experimental findings. This simplified annotation method demonstrated a consistent improvement in performance over the conventional approach, leading us to adopt it for enhancing the efficiency of our model.

For model training, we employed RoBERTa-large-PM-M3-Voc,[10] a biomedical language model pre-trained on PubMed and PMC that achieves outstanding performance on multiple BioNER tasks, as the backbone model and added two layers of MLP with ReLU activation. For the initial training phase, we integrated each BioNER dataset with a general-domain NER dataset for training. Following the multi-task learning process, we executed a subsequent fine-tuning phase exclusively on the respective BioNER datasets.

Table 3: The result of the F1 score on BioNER benchmarks.

| Target Dataset | Baselines | | | | GERBERA(Ours) | | | | |
|---|---|---|---|---|---|---|---|---|---|
| | SOTA Representatives | | Other Baselines | | General-domain NER Datasets | | | | |
| | BERN2[a] (8 BioNER datasets) | AIONER[a] (11 BioNER datasets) | PTC[b] | BioBERT[b] | CoNLL2003 | GUM | Movie | Restaurant | GUM_Time |
| BC2GM | 83.7 | N/A | 78.8 | 83.4 | 83.8 | 84.0 | 84.9 | 84.2 | **85.2** |
| NCBI-disease | 88.6 | **89.6** | 81.5 | 88.3 | 88.7 | 88.9 | 88.8 | 88.9 | 89.0 |
| Linnaeus | **92.7** | 90.6 | 85.6 | 88.0 | 81.6 | 83.0 | 87.7 | 87.4 | 84.1 |
| BC5CDR-chem | 92.6 | 92.8 | N/A | 93.2 | 93.3 | 93.4 | 93.5 | 93.5 | **94.2** |
| JNLPBA-CL | 78.6 | N/A | N/A | 75.2 | 78.7 | 78.8 | 78.6 | 78.8 | **80.7** |
| JNLPBA-CT | 80.7 | N/A | N/A | 79.4 | **82.2** | 81.2 | 82.1 | 81.6 | **82.2** |
| JNLPBA-RNA | 76.5 | N/A | N/A | 77.4 | 79.5 | 78.9 | 79.6 | 79.7 | **82.1** |
| JNLPBA-DNA | 77.8 | N/A | N/A | 76.8 | 80.3 | 80.0 | **80.9** | 80.2 | 80.5 |
| Average | 83.9 | N/A | N/A | 82.7 | 83.5 | 83.5 | 84.5 | 84.3 | **84.8** |

*Note:* The results of the GERBERA method trained using the target BioNER dataset and one general-domain NER dataset. "CoNLL2003," "GUM," "Movie," "Restaurant," and "GUM_Time" indicate the general-domain NER datasets used during multi-task learning. "Movie" and "Restaurant" represent the MIT_movie and MIT_restaurant datasets, respectively. The "GUM_Time" dataset is formed by extracting time annotations from the GUM dataset. The highest F1 scores achieved are highlighted in bold.

---

[10] https://dl.fbaipublicfiles.com/biolm/RoBERTa-base-PM-M3-Voc-hf.tar.gz

**a**: The results of BERN2 [24] and AIONER [45] are directly sourced from their publications. Both models were trained using multiple BioNER datasets—eight for BERN2 and eleven for AIONER. For considerations of dataset size and training efficiency, we have opted to use the BC5CDR-chemical [36] dataset in place of the BC4CHEMD [48] dataset used in the BERN2 study.

**b:** The result of PTC [44] is derived from the BERN2 study and the results of BioBERT [13] are derived from their published experimental result.[11] Additionally, we evaluated each sub-dataset within JNLPBA [27] by fine-tuning the BioBERT model using three different random seeds, and the average results were calculated.

## 3. Results

To assess the efficacy of our proposed method, we conducted evaluations on eight biomedical entity types from five BioNER datasets. We first compare the performance of our GERBERA models with baseline models to demonstrate the effectiveness of our method. Subsequently, we perform an error analysis to gain a deeper understanding of the efficacy of our method. Finally, we include ablation studies to indicate the generalization and adaptability results of our method by comparing the similarity and diversity of the introduced general-domain NER datasets.

### 3.1. Main results

The comparative performance analysis of our GERBERA model against strong baselines is presented in Table 3. Our experimental results demonstrate that the GERBERA method, with different general-domain NER datasets, consistently matches or exceeds baseline performance across six out of eight biomedical entity types, robustly and substantially enhancing the performance of the BioNER model. Notably, despite using fewer training resources, our best-performing GERBERA model trained with the GUM_Time [12] dataset exhibits a competitive edge even when compared to strong baselines: multi-task learning with eight BioNER datasets (i.e., BERN2) and training with eleven BioNER datasets (i.e., AIONER). Our model achieved substantial improvements in F1 scores for datasets with limited training instances, with JNLPBA-CL improved by 2.1%, JNLPBA-DNA by 2.7%, and JNLPBA-RNA by 4.7%. While GERBERA achieves optimal performance with the GUM_Time dataset, the MIT_Movie dataset demonstrates more robust and stable improvement across all BioNER datasets. Additionally, GERBERA demonstrated relatively suboptimal performance on the Linnaeus dataset. Further analysis and discussion regarding the Linnaeus dataset are provided in the Discussion section of our study.

---

[11] https://github.com/dmis-lab/biobert-pytorch/tree/master/named-entity-recognition.
[12] The GUM_Time dataset, formed by extracting time annotations from the GUM dataset, is explained in detail in the Ablation study section.

## 3.2. In-depth analysis

To more thoroughly understand the efficacy of our method, we conducted an error analysis based on the GERBERA model training with the GUM_Time dataset. The analysis indicated that our approach consistently and significantly decreased both false positive and false negative predictions across all BioNER datasets except the Linnaeus dataset. Specifically, GERBERA achieved an average reduction of 13.8% in false positives and 12.5% in false negatives compared to the baselines. Notably, GERBERA also demonstrated higher accuracy in identifying entity boundaries across all BioNER datasets, reducing boundary mismatch cases by an average of 15.6%. This enhancement enabled more precise determination of the start and end positions of various biomedical named entities, reducing errors due to unclear boundaries. For instance, while the baseline incorrectly detected the boundaries and identified entities "[35S] TBPS" and "crack cocaine" as two separate entities—"[35S]" and "TBPS", "crack" and "cocaine"—GERBERA correctly recognized them as single entities.

Additionally, we evaluated the generalizability of GERBERA method using three types of recognition ability metrics [49]: (1) Memorization, the ability to identify entity mentions seen during training; (2) Synonym generalization, the ability to recognize new surface forms of existing entities; and (3) Concept generalization, the ability to identify novel entity mentions or concepts not previously encountered. The results indicate that GERBERA outperformed the BERN2 baseline, with an average improvement of 1.1% in memorization, 0.8% in synonyms, and 1.4% in new concepts on the NCBI-disease and BC5CDR-chemical datasets. Notably, GERBERA achieved 98.4% in memorization (BERN2: 97.3%), 85.7% in synonym generalization (BERN2: 81.6%), and 89.2% (BERN2: 86.3) in concept generalization on the BC5CDR-chemical corpus, highlighting the strong generalizability. These enhancements expand the recognition range of the BioNER model, enabling the effective identification of rare or novel entity tokens.

Table 4: Ablation study of entity types in the GERBERA framework using CoNLL2003.

| General | JNLPBA-RNA | JNLPBA-CT | NCBI-disease | BC5CDR-chem | Average |
|---|---|---|---|---|---|
| N/A | 78.7 | 81.2 | 88.1 | 93.4 | 85.4 |
| CoNLL2003-LOC | 79.2 | 81.2 | 88.2 | 93.4 | 85.5 |
| CoNLL2003-PER | 78.6 | 81.5 | 88.1 | **93.5** | 85.4 |
| CoNLL2003-ORG | **80.0** | 81.8 | **88.8** | 93.1 | 85.9 |
| CoNLL2003 | 79.5 | **82.2** | 88.7 | 93.4 | **86.0** |

*Note*: N/A means the model is exclusively fine-tuned on the target dataset. The bold numbers indicate the highest F1 scores.

## 3.3. Ablation study

Given that BioNER datasets typically focus on a single entity type, we sampled entity types without replacement from each general-domain NER dataset to create diverse sub-datasets. We investigated the influence of individual entity types in general-domain NER datasets on the performance of BioNER models. We divided the CoNLL2003 dataset into three separate sub-datasets according to entity type: LOC, PER, and ORG.[13] Each sub-dataset contained the same set of training sentences, but with single entity-type labels. We used JNLPBA-RNA, JNLPBA-CT, NCBI-disease, and BC5CDR-chemical datasets as the target datasets. The evaluation results, presented in Table 4, reveal that different general-domain entity types have varying effects on BioNER model training even from the same dataset. In addition, multi-task learning with the same general-domain entity type with different biomedical entity types also has different effects on model performance. While using all entity types did not always yield optimal performance for each target dataset, it was generally a more reliable choice for enhancing overall performance.

Table 5: The performance of GERBERA models with different general-domain NER datasets.

| Dataset | BERN2 | GERBERA Variants | | | | |
| --- | --- | --- | --- | --- | --- | --- |
| | | ORG[1] | ACTOR[2] | DISH[3] | TIME[4] | ALL[5] |
| BC2GM | 83.7 | 84.4 | 85.0 | 84.4 | **85.2** | 84.5 |
| NCBI-disease | 88.6 | 88.8 | 89.0 | 88.8 | **89.0** | 88.7 |
| Linnaeus | **92.7** | 86.5 | 87.5 | 86.3 | 84.1 | 83.3 |
| BC5CDR-chem | 92.6 | 93.1 | 93.7 | 93.4 | **94.2** | 93.5 |
| JNLPBA-CL | 78.6 | 79.0 | 79.4 | 79.7 | **80.7** | 79.3 |
| JNLPBA-CT | 80.7 | 82.0 | 81.9 | 81.5 | **82.2** | 81.5 |
| JNLPBA-RNA | 76.5 | 80.0 | 80.7 | 81.2 | **82.1** | 78.5 |
| JNLPBA-DNA | 77.8 | 79.2 | 78.8 | 80.2 | **80.5** | 80.3 |
| Average | 83.9 | 84.2 | 84.5 | 84.4 | **85.2** | 83.7 |

*Note*: **ORG**[1] denotes training with the CoNLL2003-ORG dataset; **ACTOR**[2] involves training with the MIT_Movie-Actor dataset; **DISH**[3] denotes training with the MIT_Restaurant-Dish dataset; **TIME**[4] denotes training with the GUM-Time dataset; **ALL**[5] denotes training with all these sub-datasets. Bold numbers indicate best F1 scores.

In a further exploration of the effect of different entity types from different general-domain datasets on BioNER datasets, we employed four distinct general-domain sub-datasets: (1) CoNLL2003-ORG: the CoNLL2003 dataset containing only 'Organization' annotations; (2) GUM-Time: the GUM dataset containing only 'Time' annotations; (3) MIT_Movie-Actor: the MIT_Movie dataset containing only 'Actor' annotations; and (4) MIT_Restaurant-Dish: the MIT_Restaurant dataset containing only 'Dish' annotations. Furthermore, to assess whether there are benefits to learning from a broader and more diverse set of contexts, we merged all these general-domain sub-datasets simultaneously during training. Table 5 indicates general-domain NER sub-datasets led to consistent positive effects on model performance on all BioNER datasets except the Linnaeus dataset, surpassing the performance of the baseline model. Notably, the results indicate that training with merged multiple general-domain NER datasets did not accumulate the positive effect of each general-domain NER dataset. Additionally, our GERBERA method, when trained with the GUM_Time dataset, achieved optimal performance across most BioNER datasets except the Linnaeus dataset (see Discussion for further details).

---

[13] We retain only one annotation type while labeling all other types as "O" for all general-domain NER sub-datasets in our study, following the division method of the BC5CDR [36] dataset.

## 4. Discussion

Recent multi-task BioNER models, which were trained using various BioNER datasets, have demonstrated superior performance compared to models trained on a single BioNER dataset [28-31, 50]. These studies all focused on leveraging multiple biomedical datasets to improve BioNER performance. However, as we discussed in Section 1, acquiring large-scale annotated biomedical datasets is typically costly, and the performance of BioNER models on small-scale biomedical datasets is considerably inferior to the performance of models on large-scale BioNER datasets. Additionally, multi-task models trained on multiple BioNER datasets struggle to achieve optimal performance on each dataset simultaneously. Moreover, utilizing multiple BioNER datasets for training could easily introduce label ambiguity stemming from overlapping concepts across different biomedical corpora.

Our study presents several key findings to tackle these challenges. First, the introduction of general-domain NER datasets enhances the BioNER model performance, where our model achieves outstanding performance on most biomedical datasets. Additionally, the introduction of general-domain NER datasets can significantly reduce the reliance of BioNER models on large-scale gold standard corpora, where our models improve even more on small-scale datasets, which highlights the efficiency of our approach on limited-size biomedical datasets. Furthermore, our method is robust to the introduction of different general-domain NER datasets and consistently improves BioNER model performance across six out of eight biomedical entity types, regardless of the specific general-domain NER datasets or entity types used. Therefore, GERBERA is a reliable method to enhance the performance of the BioNER model using low-cost general-domain NER datasets. This method not only demonstrates notable and robust performance on most BioNER datasets, but also substantially reduces the dependence of BioNER models on large-scale gold standard datasets.

The primary limitation of our method is the suboptimal performance on the Linnaeus dataset. The Linnaeus dataset is noted for its limited diversity and frequent repetitiveness of organism names, which often leads to overfitting of the model and overestimation of the performance [51]. Moreover, the inherent issues of limited diversity and label inconsistency within the Linnaeus dataset pose severe problems for training biomedical language models [40, 52]. In our additional experiments, fine-tuning exclusively on the single Linnaeus dataset outperformed multi-task learning with a general-domain NER dataset. This outcome contradicted the performance of multi-task learning models trained on various BioNER datasets, which typically show excellent performance on the Linnaeus dataset. We found that GERBERA consistently reduced the false positives and outperformed the baseline in terms of precision, achieving a precision of 5.6% higher than that of the BERN2 [24] baseline, which aligns with the performance across other BioNER datasets. However, GERBERA exhibited a considerably lower recall value on the Linnaeus dataset—12.1% less than the baseline—with a remarkable increase in false negatives, resulting in a diminished F1 score. Further investigation into the significant reduction in recall values can be attributed to the label ambiguity of the concept overlap between the species entity type recognized in the Linnaeus dataset and those in the general-domain NER dataset. For instance, entities such as "people", "patient", and "human" are labeled as "B" in the Linnaeus dataset but are designated as "O" in the CoNLL2003 dataset. This discrepancy can confuse the BioNER model, leading to missing matches.

Another limitation of the GERBERA method, compared to multi-task learning BioNER models such as BERN2, is the decreased efficiency when inferring multiple biomedical entity types simultaneously. For example, BERN2 has approximately 365 million parameters in a single model that supports the efficient inference of multiple entity types. In contrast, our method requires a separate BioNER model for each target biomedical entity type with roughly 357 million parameters, resulting in decreased inference speed and heightened disk usage requirements. However, we would like to emphasize that our focus was on developing a more effective learning framework in terms of performance without relying on expensive domain-specific resources. Additionally, augmenting BioNER models using general-domain NER datasets represents a novel approach not previously introduced in this domain. In future research, we will aim to effectively combine general-domain resources with existing multi-task models that infer multiple entity types at once to boost the performance while maintaining inference efficiency.

## 5. Conclusion

This paper presented a new transfer learning method that leverages accessible and cost-effective general-domain NER datasets to augment BioNER tasks. Through extensive testing across five BioNER datasets, our method not only demonstrated the potential to leverage a broad range of fundamental linguistic features from a general-domain NER dataset to improve the overall understanding of biomedical entity recognition but also surpassed the performance of benchmarks. Notably, our method exhibited exceptional strength on datasets with limited availability. Additionally, a notable aspect of our approach lay in its robust generalization capability, as evidenced by consistent performance improvements across various general-domain datasets introduced. This highlights the versatility and efficacy of our method, making it a valuable asset for BioNER tasks, especially in scenarios where biomedical datasets are scarce or costly to obtain. We hope that our method, codes, and models will make a valuable asset for scenarios with scarce or costly biomedical datasets and facilitate further research in this direction.

## 6. Ethics statement

This paper is not concerned with human evaluation. The authors foresee no potential ethical issues with the work presented in this paper.

# 7. Acknowledgement

This research was supported by 1K99LM014024, 1R01AG078154, and the NIH Intramural Research Program (IRP), National Library of Medicine.